\documentclass[sigconf,nonacm]{acmart}
\AtBeginDocument{%
  }

\setcopyright{none}
\renewcommand\footnotetextcopyrightpermission[1]{}
\begin{document}

\title{A Composition-Aware Pretraining Framework for Geospatial Foundation Models}

\author{Aryan Kashyap Naveen}
\email{aryankashyapnaveen@gmail.com}
\orcid{1234-5678-9012}
\affiliation{%
  \institution{National Institute of Technology, Karnataka}
  \city{Surathkal}
  \state{Karnataka}
  \country{India}
}

\author{Abhishek Srinivas}
\email{sriabhi2407@gmail.com}
\affiliation{
  \institution{Visa Inc}
  \city{Bangalore}
  \state{Karnataka}
  \country{India}
}

\author{Pranav Moothedath}
\email{pranavbijum@gmail.com}
\affiliation{%
  \institution{Greenlight}
  \city{Bangalore}
  \state{Karnataka}
  \country{India}
}

\author{Shrutilipi Bhattacharjee}
\email{shrutilipi@nitk.edu.in}
\affiliation{%
  \institution{National Institute of Technology, Karnataka}
  \city{Surathkal}
  \state{Karnataka}
  \country{India}
}




\renewcommand{\shortauthors}{Naveen et al.}

\begin{abstract}
Geospatial and remote sensing technologies provide an unprecedented volume of data for earth observation, driving critical applications such as zero-shot image retrieval, object detection, land cover classification, and change detection. Recently, geospatial foundation models have emerged as state-of-the-art methods for these tasks. However, existing pretraining methodologies, predominantly based on patch reconstruction via masked autoencoders or global contrastive learning, process imagery through a single-concept lens. This aggregation fails to capture the highly compositional nature of complex satellite scenery. We propose a composition-aware pretraining framework that explicitly encodes these fractional land-cover mixtures. An offline visual vocabulary of fundamental geospatial textures is constructed by clustering patch embeddings extracted from a frozen vision transformer. Each satellite image cell is then mapped to a histogram denoting its exact fractional makeup, a representation we term the ``composition target''. These targets serve as the primary prediction objective, distilled into the backbone via the Sinkhorn Earth Mover's Distance, which penalizes predictions in proportion to the semantic distance of their error. A Bidirectional Multi-Instance Learning (MIL) objective provides complementary categorical feature alignment.

Experimental evaluation reveals that composition-aware pretraining yields substantial gains on region-level understanding tasks requiring semantic similarity judgment (zero-shot image retrieval, scene classification), while remaining competitive on tasks requiring fine-grained spatial precision (segmentation, object detection). Operating with a 36.8M parameter backbone, our framework surpasses SatMAE and Prithvi-EO-2.0 (303M and 600M parameters) on region-level understanding tasks. On the fine-grained ForestNet-12 dataset, a rigorous testbed for compositional discrimination, our method nearly doubles the baseline mAP@10 (0.279 to 0.434), directly validating the effectiveness of explicit composition modeling. The code implementation can be found \href{https://github.com/05kashyap/GFM_Composition_Pretraining}{\textbf{\textit{here}}}.
\end{abstract}



\maketitle
\section{Introduction}

\begin{figure}[h]
    \centering
    \includegraphics[width=\linewidth]{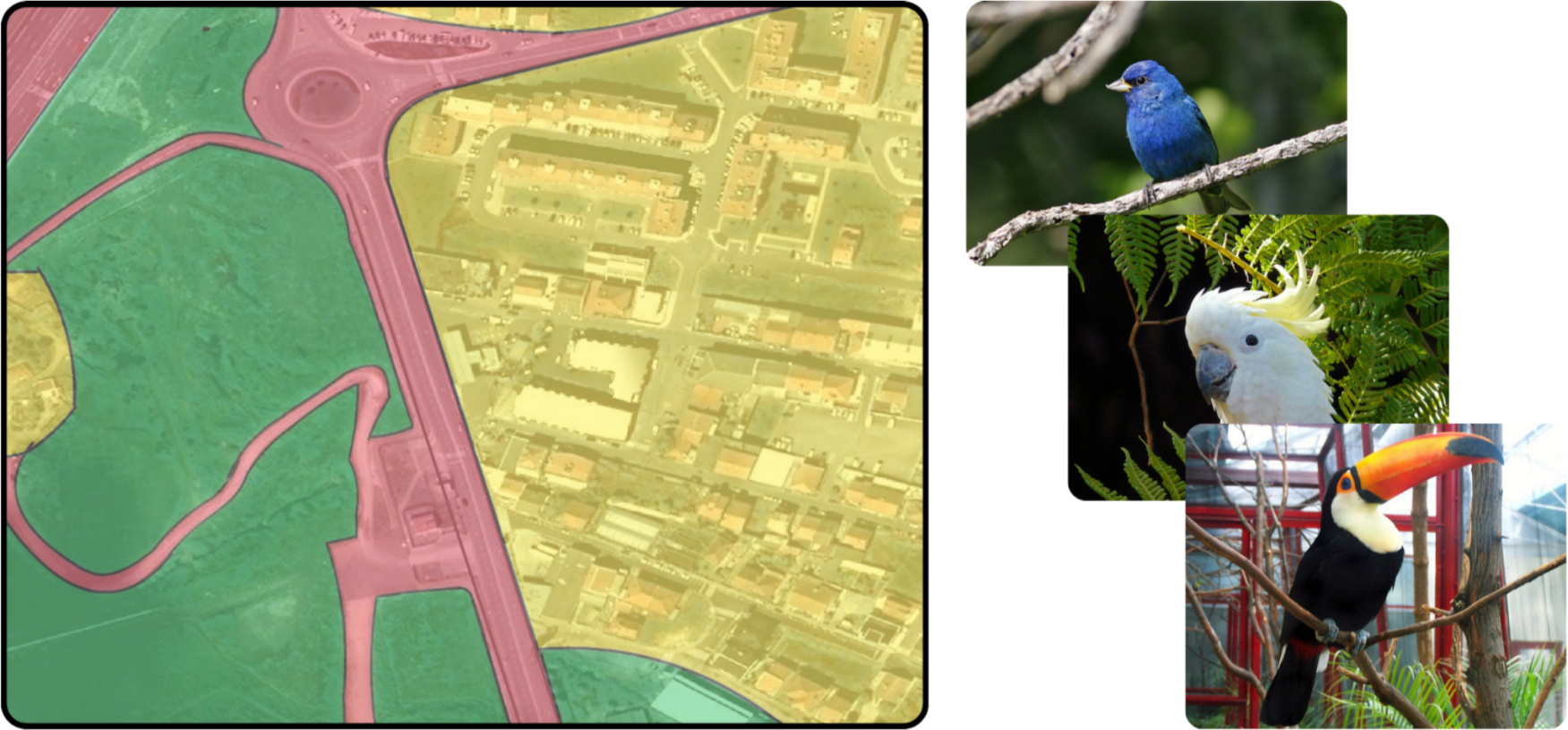}
    \caption{Contrast between the highly compositional nature of satellite imagery and object-centric datasets (e.g., ImageNet). While natural images typically feature a distinct primary subject, geospatial cells represent a mixture of multiple distinct ground-cover textures.}
    \label{fig:comparison}
\end{figure}

Satellite imagery contains rich information to improve geospatial understanding. Recently, geospatial Foundation Models (FMs) have achieved state-of-the-art performance in generating the high-dimensional embeddings required for various tasks such as zero shot image retrieval, change detection, and object detection ~\cite{chen2024dynamicvis, cong2022satmae, prithvi, guo2024skysensev2}. They are not, however, without their limitations~\cite{fm-limitations}. Geospatial FMs suffer from a fundamental limitation rooted in their pretraining paradigms: they process imagery through a single-concept lens. Unlike standard computer vision datasets, such as ImageNet, which predominantly feature single, object-centric subjects, satellite imagery is inherently compositional. A standard satellite image cell rarely contains a monolithic concept. Instead, it is a complex mixture of multiple types of land cover. A coastal port scene, for example, is a fractional mixture of calm water, concrete docks, ship hulls, road markings, and warehouse roofs. Figure \ref{fig:comparison} illustrates this dichotomy between single-subject natural images and multi-concept geospatial images.

Existing pretraining paradigms for geospatial FMs rely on patch 
reconstruction~\cite{prithvi, cong2022satmae} (masked autoencoders) or global contrastive learning~\cite{seco, guo2024skysensev2, chen2024dynamicvis}. However, these approaches fail to explicitly model the compositional nature of satellite scenery. Patch reconstruction methods force the model to waste capacity memorizing low-level and redundant background textures instead of learning semantic relationships. Meanwhile, the global contrastive approaches average heterogeneous land-cover mixtures into a single, holistic embedding. A scene containing 60\% water and 40\% buildings would produce a pooled embedding that semantically represents neither water nor buildings in isolation. Consequently, two entirely different scenes can yield nearly identical pooled vectors if their average spectral content happens to align.

In this paper, we propose a composition-aware pretraining framework that shifts the representational focus from whole cells to localized patches. We treat each image cell as a distribution over a finite vocabulary of visual textures. In our framework, first an expressive offline visual vocabulary is computed of \textit{k} fundamental geospatial building blocks by clustering high-dimensional patch embeddings extracted from a frozen self-distilled vision transformer. Every cell in the training corpus is then mapped to a target histogram using soft assignments, explicitly capturing what individual visual textures are present and in what exact proportions. We call this the "composition-target". While predicting Bag-of-Visual-Words (BoVW) representations has been explored in general computer vision \cite{obow} and classic remote sensing \cite{emd-rs}, we propose a novel integration tailored for modern geospatial foundation models. We train a highly efficient, Mamba-based backbone (DynamicVis) to predict these patch-level target histograms. By explicitly modeling fractional composition as a target, our framework enables the foundation model backbone to generate highly discriminative embeddings that effectively capture the heterogeneity of the Earth's surface. Our main contributions are summarized as:
\begin{itemize}
    \item We identify the single-concept pooling bottleneck in current geospatial foundation models and introduce a composition-aware pretraining strategy based on predicting visual texture histograms.
    \item We design a robust loss landscape combining EMD distillation with Multi-Instance Contrastive (MIL) auxiliary losses to teach a Mamba-based \cite{mamba} architecture \cite{chen2024dynamicvis} to understand semantic distances between ground-cover mixtures.
    \item The proposed method demonstrates substantial improvements across tasks that require region-level understanding. The zero-shot image retrieval and scene classification tasks see improvements of up to 56\% mAP@10 scores and 7\% accuracy, respectively.
\end{itemize}
\section{Background and Related Work} 

Remote sensing research has been fundamentally transformed by the evolution of embedding techniques and the increase in data scale. Works have transitioned from task-specific architectures~\cite{task-specific} to generalized foundation models. Foundation models have brought a notable shift in how remote sensing tasks are approached, making it possible to learn generalizable representations from large volumes of remote sensing data. Most of these models are built on transformer or state-space architectures and are pre-trained at a massive scale to capture the semantics of geospatial imagery. The resulting representations transfer well across downstream tasks, such as classification, segmentation, retrieval, and change detection, often without requiring extensive fine-tuning. The sections below trace this development across popular pretraining paradigms in the field.

\subsection{Masked Autoencoder Based Models}
The dominant pretraining paradigm for early vision foundation models heavily relied on Masked Image Modeling (MIM). As one of the pioneering foundational models for Earth observation, \textbf{Prithvi} \cite{prithvi} demonstrated the potential of this approach by pretraining a Vision Transformer (ViT) using a masked autoencoding objective on the Harmonized Landsat Sentinel-2 (HLS) dataset. Similarly, models like \textbf{SatMAE} \cite{cong2022satmae} extended this reconstructive framework to process temporal and multi-spectral satellite imagery, proving that reconstructing masked patches could yield enhanced generalizable multi-modal representations.

Despite their success in capturing generic textural patterns, MAE-based models suffer from a critical misalignment with the inherent physical properties of Earth observation imagery. Unlike object-centric natural images, satellite scenes are fundamentally characterized by massive, homogeneous backgrounds. Consequently, pixel-level reconstructive pretraining inadvertently forces the network's capacity to over-index on memorizing these redundant background statistics, rather than extracting highly discriminative, high-level semantic features for complex land-cover mixtures. While a thorough treatment of this limitation is provided by~\cite{chen2024dynamicvis, ringmo}, we briefly restate it here as it directly motivates our composition-aware pretraining objective.

\subsection{Contrastive Learning Based Models}
To circumvent the limitations of pixel reconstruction, researchers explored contrastive learning approaches, which optimize the latent space by pulling semantically similar views together while pushing dissimilar ones apart. Early adaptations, such as \textbf{SeCo}~\cite{seco}, \textbf{SkySense}~\cite{guo2024skysensev2} and \textbf{RS-M-CLIP}~\cite{multiling} focused on aligning global, image-level representations. However, this coarse semantic alignment tends to dilute localized, fine-grained details, effectively collapsing complex, multi-class spatial compositions into a single, diluted holistic embedding. Addressing the severe target sparsity and spatial redundancy of remote sensing, \textbf{DynamicVis} \cite{chen2024dynamicvis} introduced a highly efficient state-space model backbone governed by an adaptive token routing mechanism. To train this dynamic router, the authors proposed a Region-Level Meta-Embedding Multi-Instance Learning (MIL) pretraining paradigm. By treating high-resolution scenes as "bags" of instances and aligning regional visual embeddings with language-guided categorical meta-embeddings, DynamicVis successfully learnt to decouple heterogeneous foreground instances from expansive backgrounds. While the bidirectional MIL approach in DynamicVis provides categorical separation, it still processes regions through a discrete, single-concept classification lens. It does not explicitly quantify the proportional distribution of land-cover categories within a complex patch.

\section{Methodology}

This section details the proposed composition-aware pretraining framework. First, we explore the creation of a patch-level pseudo dataset; then we discuss the generation of quantitative composition targets, and finally, the geospatial foundation model pretraining framework is explained.

\subsection{Pretraining Dataset Creation}

To capture the high variance of textures within a single geospatial scene, we shift the representational focus from whole images to localized visual patches. We first process the large-scale pretraining dataset containing raw satellite imagery by dividing it into standard $512 \times 512$ pixel cells. Each patch of these cells is embedded using a ViT encoder to generate expressive descriptions for the composition of each cell. Then the embeddings are clustered, assigned to a visual vocabulary, and soft-assigned per cell to generate composition targets. 

\subsubsection{Patch Representation}

\begin{figure}[t]
    \centering
    \includegraphics[width=\linewidth]{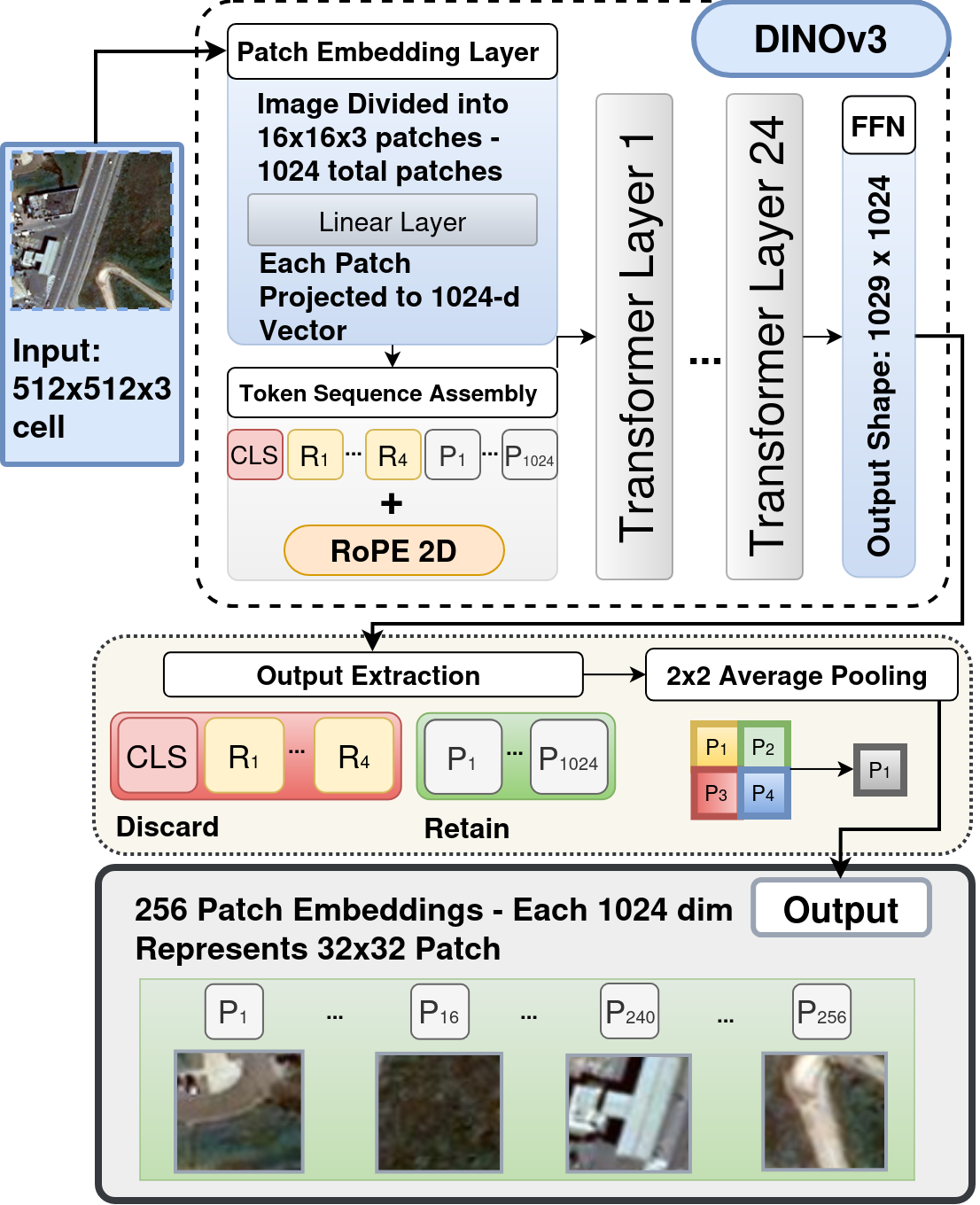}
    \caption{Generate embeddings for each patch in a satellite image cell utilizing average-pooling.}
    \label{fig:dataset_patch_embed}
\end{figure}

Each cell is passed through a frozen, pretrained DINOv3 (ViT-L/16) Vision Transformer \cite{dinov3} to extract dense representations (Figure \ref{fig:dataset_patch_embed}). Internally, the DINOv3 architecture natively patches the $512 \times 512$ cell into a grid of $16 \times 16$ pixel patches. This results in 1024 patches; each of these patches is projected to a 1024 dimension spatial patch token. During the forward pass, these tokens are processed alongside a class (CLS) token and four register tokens. The register tokens serve as a scratch space, absorbing high-norm global artifacts that would otherwise pollute the spatial representations. To maintain strict spatial awareness throughout the forward pass, 2D Rotary Position Embeddings (RoPE) are added to the token sequence. 

The spatial tokens pass through the 24 transformer layers. The self-attention mechanism ensures that every token attends to every other token in the cell. Consequently, the resulting feature vectors are context-aware embeddings. For example, the token for a patch of water inherently encodes its surrounding semantic context. This could help distinguish open ocean from water adjacent to a harbor. After the transformer layers, the intermediate outputs are passed through a feed forward network (FFN); the final output dimension is $1029 \times 1024$ (1029 tokens of 1024 dimensions). The CLS and the four register tokens are discarded, leaving only the 1024 contextualized spatial patch tokens.

To improve computational efficiency and reduce storage overhead during the subsequent clustering phase, we apply a spatial downsampling step. The 1024 patch tokens are average-pooled in $2 \times 2$ spatial groups. This yields 256 dense embeddings per cell, where each resulting token effectively represents a $32 \times 32$ pixel region. This approach preserves the discriminative spatial semantics of the scene while significantly compressing the data, producing highly expressive and context-aware patch token embeddings across the pretraining corpus.

\subsubsection{Composition Targets Generation}

\begin{figure}[t]
    \centering
    \includegraphics[width=\linewidth]{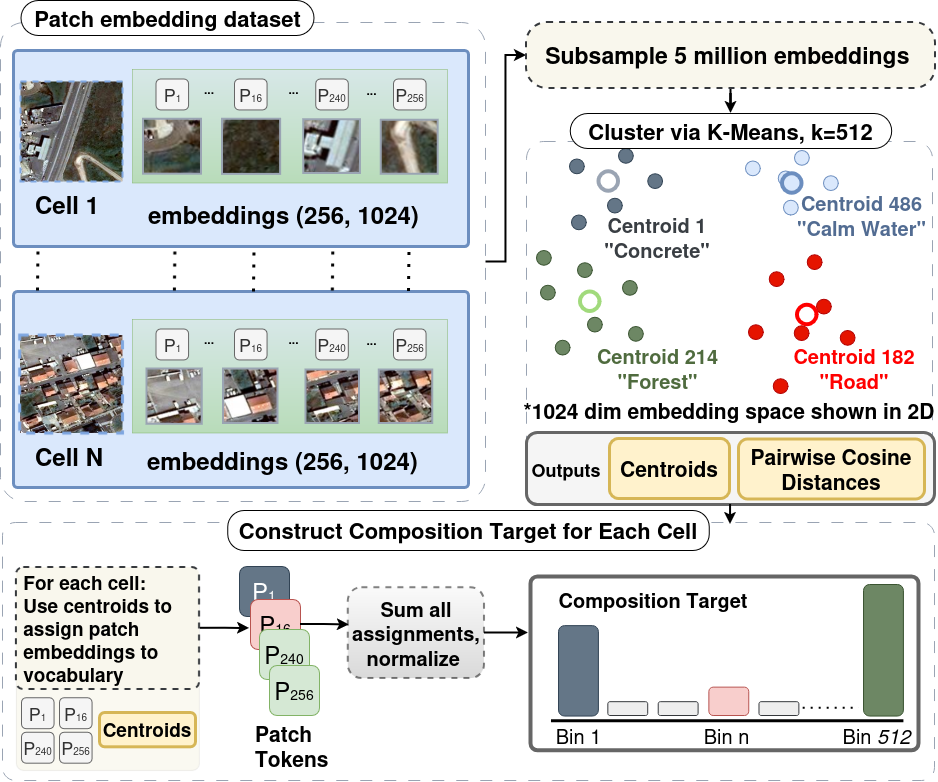}
    \caption{The patch embeddings are clustered to create a visual vocabulary. Then the patch tokens for each cell are assigned to the vocabulary. All assignments are summed and normalized to generate the final composition target.}
    \label{fig:dataset}
\end{figure}

The patch embeddings generated in the previous step are used to quantify the compositional makeup of each cell (Figure \ref{fig:dataset}). To achieve this, each patch must be mapped to a ``concept'' or ``visual texture''. We perform K-Means clustering on the extracted patch embeddings to discover the fundamental visual texture of our dataset, defining $k=512$ distinct concepts ($C_1 - C_k$). The earlier collected patch embeddings were roughly 93 million individual 1024-dimensional vectors across the full pretraining dataset (363,000 cells × 256 embeddings each).

Clustering all 93 million embeddings simultaneously was computationally infeasible. Instead, a random subsample of 5 million embeddings was drawn uniformly across cells. 
Before clustering, every token is $L_2$-normalized. Without normalization, K-Means would cluster by vector magnitude rather than by semantic direction, and tokens representing ``dark water'' and ``bright water'' would end up in different clusters despite being semantically similar simply because one has higher pixel intensity. The normalized 5 million tokens are clustered using FAISS GPU-accelerated K-Means with K=512. These 512 cluster centroids form our global visual vocabulary, where each centroid represents a recurring and meaningful geospatial texture. After clustering, a $512 \times 512$ ground cost matrix is precomputed. The matrix encodes how semantically similar every pair of vocabulary textures is via cosine similarity. This matrix is used later during training as the geometric substrate for the Earth Mover's Distance loss \cite{sinkhorn}. 

To compute the histogram targets, we start by iterating over each cell; the 256 patch tokens are loaded and $L_2$-normalized. For each token, its cosine similarity to every one of the 512 centroids is computed as their dot product. Rather than assigning each token to its single nearest centroid, we use a soft assignment. The assignment weight of token a to centroid is computed using a Radial Basis Function (RBF) kernel. The effect is that each token distributes its probability mass across multiple nearby centroids rather than committing entirely to one. 
We opt for soft rather than hard assignment because deep embedding spaces are continuous and forcing a binary boundary between semantically adjacent clusters introduces a discretization error that could propagate into the training signal. 

Once every patch token in the cell has been soft-assigned to the 512 centroids, the assignments are aggregated into a single histogram. For each centroid k, all 256 per-patch weights are summed. This gives a 512-dimensional vector where each entry represents the total soft-assignment mass that flowed to that centroid across all patches in the cell. Finally, the vector is then $L_1$-normalized. Ultimately, this $L_1$-normalized histogram we term the "composition target", provides a precise and fractional description of the fundamental visual textures that construct the scene.

\begin{figure*}[t]
    \centering
    \includegraphics[width=\textwidth]{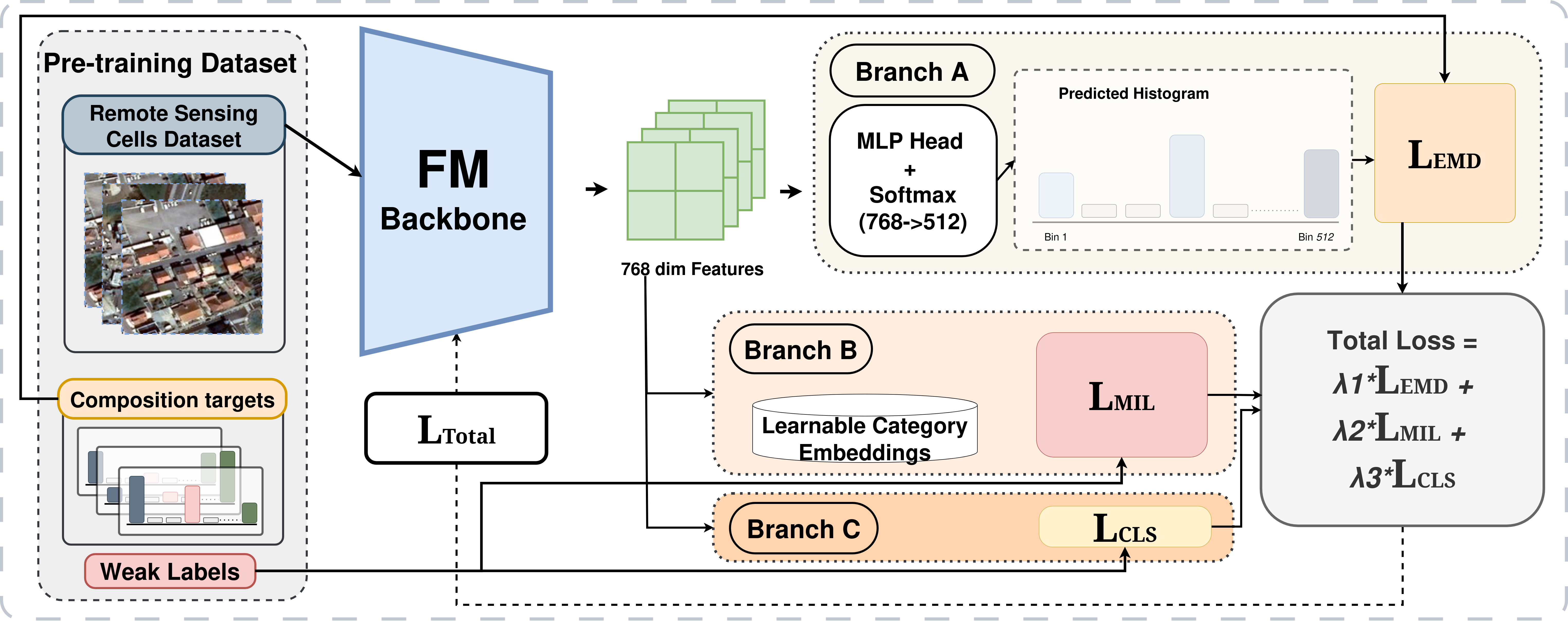}
    \caption{The proposed Composition-Aware Pretraining Architecture. The input satellite cell is processed by the DynamicVis backbone to extract a dense feature representation. This feature tensor is then routed through three parallel objectives: (A) A BoVW head predicting a fractional texture histogram optimized via Sinkhorn EMD, (B) a Bidirectional MIL branch aligning raw features with learnable category embeddings, and (C) an Auxiliary classification head operating on detached features for stabilization.}
    \label{fig:architecture}
\end{figure*}
\subsection{Foundation Model Training}

During the online pretraining phase (Figure \ref{fig:architecture}), the objective is to teach an efficient foundation model (FM) backbone to intrinsically understand both the composition and semantic categorization of an input scene. The training utilizes three inputs: the raw pre-training dataset images, the offline-generated histogram targets, and weak image-level annotations.

\subsubsection{Backbone Architecture}

To process the complex, multi-scale nature of remote sensing imagery, we employ the DynamicVis-base architecture \cite{chen2024dynamicvis} as our foundation model backbone. DynamicVis is available in two variants, Base (36.8M parameters) and Large (91.3M parameters). Both variants process 3-channel RGB imagery. We adopt the Base variant to prioritize parameter efficiency. Unlike traditional Vision Transformers that rely on uniform dense processing and allocate equal computational resources across vast, redundant backgrounds, DynamicVis introduces a Dynamic Region-Aware State Space Model (SSM) explicitly tailored for spatial sparsity. The architecture utilizes a Spatial-Preserving Hierarchical Patch Merger. Instead of employing aggressive patch embeddings that irreversibly compromise fine-grained details, this module uses a progressive, small-stride down-sampling strategy across hierarchical stages to strictly preserve the granular geometric configurations necessary for resolving tiny targets. To efficiently process the resulting high-resolution token sequences without immense computational waste, the backbone integrates a Dynamic Sparse Mixer composed of Adaptive Token Routing and Incremental Modeling (ATRIM) units. The ATRIM unit operates through three sequential mechanisms.

The module partitions the input sequence by isolating dense global semantics through 1D pooling, while simultaneously utilizing a top-$k$ routing mechanism to explicitly identify a sparse subset of highly salient regional tokens. The pooled global context and the selected salient regional tokens are concatenated and processed through a streamlined, bidirectional (forward and reverse) Mamba~\cite{mamba} block. This focuses the intensive $O(N)$ state-space computation exclusively on the most informative spatial elements. Crucially, to prevent the fragmentation of spatial topology, unselected background tokens are not permanently discarded. They bypass the heavy SSM computation and are reintegrated into the sequence via parameter-free residual connections, preserving the macroscopic structural continuity necessary for pixel-level interpretation. The input cell is passed through the FM Backbone, which outputs a continuous 768-dimensional latent feature tensor. The embedding is simultaneously routed through three distinct parallel branches.

\subsubsection{Branch A: Histogram Distillation via Sinkhorn EMD}
The latent feature is passed through a Multi-Layer Perceptron (MLP) head and a softmax activation function to project the 768-dimensional embedding into a 512-bin predicted probability distribution, denoted as $p$. This prediction is compared against the ground-truth Composition Target histogram, denoted as $q$. We utilize the Earth Mover's Distance (EMD), or Wasserstein-1 distance, rather than standard Mean Squared Error. Given a cost matrix $C \in \mathbb{R}^{k \times k}$, where $C_{i,j}$ represents the pairwise semantic distance between vocabulary centroids $C_i$ and $C_j$, the EMD objective is formulated as finding the optimal transport plan $T$:

\begin{equation}
\mathcal{L}_{EMD}=\min_{T \ge 0,\, T\mathbf{1}=p,\, T^T\mathbf{1}=q} 
\sum_{i=1}^k \sum_{j=1}^k T_{i,j} C_{i,j}
\end{equation}
We approximate this using the Sinkhorn-Knopp algorithm \cite{sinkhorn} to ensure computational efficiency during training. The optimal transport problem is transformed into a strictly convex formulation solvable via highly parallelizable, iterative matrix multiplications. The approximation scales efficiently on GPU hardware while providing the continuous gradients necessary for backpropagation. Because this loss incorporates the ground distance $C$ between vocabulary clusters, it penalizes the network proportionally to the severity of its error. This allows it to differentiate between mild misclassifications and severe ones.
\subsubsection{Branch B: Feature Alignment via Bidirectional MIL} 
The parallel branch processes the raw 768-dimensional embedding and routes it to a Multi-Instance Learning (MIL) branch to explicitly structure the latent space. Utilizing the million-scale dataset with its weak, region-level annotations, we treat each massive image as a ``bag'' of instances. The model explicitly learns to decouple heterogeneous foregrounds from the background by contrasting region specific visual representations against categorical meta-embeddings. To achieve this, we apply the Multi-Instance Learning Noise Contrastive Estimation (MIL-NCE) loss, formulated as:
\begin{equation}
\mathcal{L}_{MIL-NCE}=-\log \frac{\sum_{(v,t) \in P} f(v, t)}{\sum_{(v,t) \in P} f(v, t) + \sum_{(v',t') \in N} f(v', t')}
\end{equation}
where $f(x, y)=\exp(\langle x, y \rangle / \tau)$. Here, $\langle x, y \rangle$ denotes cosine similarity, $P$ represents a set of positive matches between regional visual embeddings ($v$) and their corresponding categorical meta-embeddings ($t$), and $N$ comprises negative pairs sampled from non-associated categories within the mini-batch. The parameter $\tau$ acts as a learnable temperature scalar. This bidirectional MIL loss ensures that the feature is pulled toward its correct category embedding, and conversely, that the category embedding is pulled toward all batch instances sharing that label.

\subsubsection{Branch C: Auxiliary Classification (CLS)}
To further stabilize the training dynamics and provide a grounded discriminative signal, we implement a standard auxiliary classification branch. Crucially, before the latent feature $f$ enters the auxiliary classification head, we apply a stop-gradient (detach) operation. This ensures that the classification loss only trains the auxiliary mapping weights and does not backpropagate gradients into the DynamicVis backbone, preventing the single-label classification objective from dominating or overwriting the fine-grained compositional features learned by the EMD and MIL branches.

\subsubsection{Total Objective Formulation}
The complete pretraining objective is a weighted sum of the compositional distillation, feature alignment, and stabilization branches. The total loss $\mathcal{L}_{total}$ is formulated as:$$\mathcal{L}_{total} = \lambda_{emd} \cdot \mathcal{L}_{emd} + \lambda_{mil} \cdot \mathcal{L}_{mil} + \lambda_{cls} \cdot \mathcal{L}_{cls}$$ This weighting scheme is designed to prioritize the fine-grained compositional EMD loss while allowing the MIL and CLS losses to gently constrain the global topology of the embedding space. The empirical justification for the chosen values of $\lambda_{emd}$, $\lambda_{mil}$, and $\lambda_{cls}$ is provided in Section~\ref{sec:loss_ablation}.

\section{Results and Analysis}
We evaluate the proposed framework across region-level understanding, instance-level perception, and pixel-level dense prediction tasks, preceded by an analysis of the pretraining pipeline and followed by ablation studies on vocabulary size and loss formulation. For all downstream evaluations, backbone weights are kept strictly frozen and only lightweight task-specific heads are trained, ensuring that reported performance reflects representational quality rather than fine-tuning capacity.
\begin{figure*}[h]
    \centering
    \includegraphics[width=\textwidth]{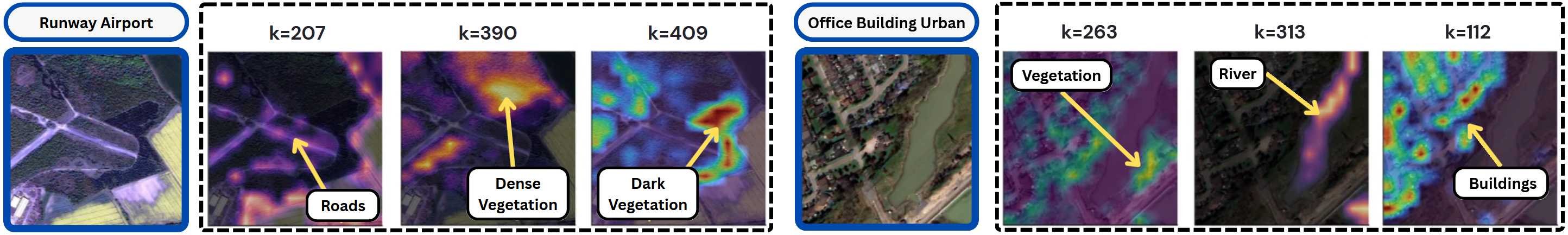}
    \caption{Spatial attention diversity across multiple cells. Overlaying the top diverse centroid activations reveals that specific clusters focus on distinct, localized geographical features.}
    \label{fig:spatial_heatmaps}
\end{figure*}

\subsection{Pretraining Pipeline Analysis}
Before evaluating on downstream tasks, we first analyze the internal dynamics of our composition-aware pretraining pipeline to validate the visual vocabulary and training stability.
\subsubsection{Pretraining Dataset}
We utilize the Functional Map of the World (fMoW) dataset in our pretraining framework. fMoW is a large-scale remote sensing dataset designed to facilitate the understanding of complex spatial patterns across the globe. It comprises more than one million high-resolution satellite images spanning more than 200 countries. The dataset encompasses a highly diverse range of human-made structures, land-use patterns, and natural environments, categorized into 62 classes.  

To adapt this raw imagery for our patch-level process, the data is processed into a uniform grid structure. First, the imagery is tiled into cells of size $512 \times 512$ pixels. These cells serve as the raw inputs for the DynamicVis backbone. Next, to explicitly capture the compositional information within each scene, every $512 \times 512$ cell is further subdivided into 256 individual $32 \times 32$ pixel patches. This decomposition ensures that the massive variance of localized geospatial textures present within the satellite imagery is physically isolated. These small patches are embedded using DINOv3 to construct the global visual vocabulary and generate the fractional histogram targets that drive our composition-aware pretraining objective.
\subsubsection{Composition Target Generation}
The foundation of our approach relies on extracting a robust, offline visual vocabulary. By embedding individual $32 \times 32$ patches using a frozen DINOv3 encoder, we captured a highly expressive latent space of localized textures. Applying K-Means clustering ($K=512$) to these patch embeddings successfully grouped the visual primitives into distinct semantic concepts. Figure \ref{fig:spatial_heatmaps} highlights the spatial attention diversity inherent in our fractional targets. The figure overlays the top diverse centroid activations onto the original cells; it is evident that individual clusters accurately ground themselves to specific regions and features within the image. Consequently, scenes with disparate land-cover mixtures produce orthogonal probability distributions, providing a rich, localized, and high-variance target for the backbone to learn from.

\subsubsection{Foundation Model Training}
We trained the DynamicVis backbone till convergence on the fMoW dataset with a batch size of 256 across two Nvidia H100 GPUs. We found that the multi-branch loss framework was necessary for stability, as the Sinkhorn Earth Mover's Distance (EMD) loss smoothly converged with the presence of Multi-Instance Learning (MIL) and Classification (CLS) losses. To evaluate the out-of-the-box performance of the trained foundation model, we apply it to various downstream tasks detailed below.


\subsection{Evaluation on Region-Level Understanding}
\begin{figure}[t]
    \centering
    \includegraphics[width=\linewidth]{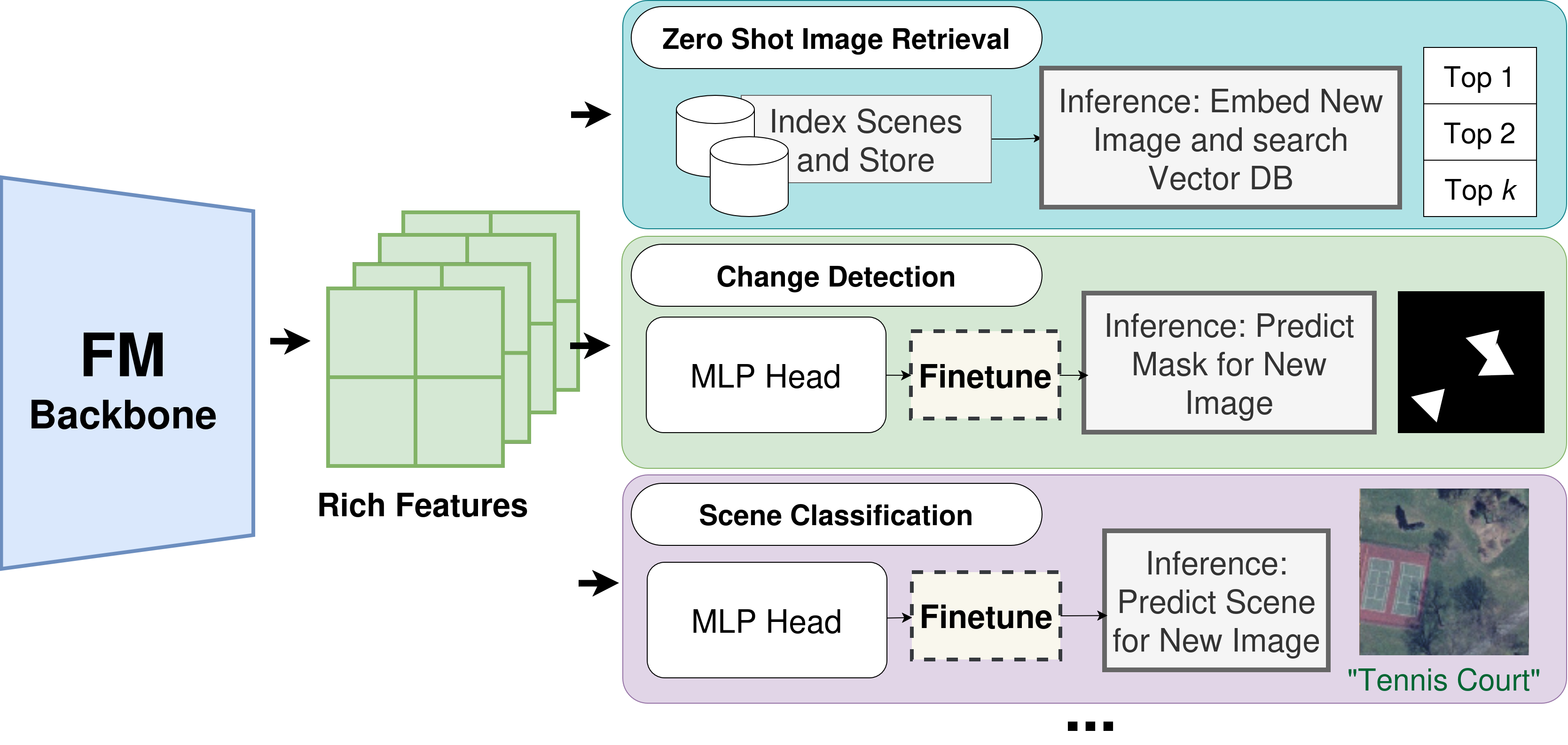}
    \caption{The FM can be applied to a variety of downstream tasks out of the box by utilizing the backbone feature maps.}
    \label{fig:downstream}
\end{figure}
For each downstream task, we compare our proposed composition-aware approach against the vanilla DynamicVis base model, the large variant, and the large-scale SatMAE and Prithvi-EO-2.0 foundation models. To strictly evaluate the quality of the learned representations, the foundation model backbones were kept completely frozen during all downstream task evaluations. Only the lightweight, task-specific heads were trained (Figure \ref{fig:downstream}).

Region-level understanding encompasses image classification, region classification, and zero-shot image retrieval. For image classification, category logits are generated by applying average pooling to the highest-resolution semantic feature map, followed by a linear projection layer. Scene classification is achieved by extracting the pooled features and passing them through a lightweight classification head. Zero-shot image retrieval involves using the high-level semantic features extracted by the backbone to retrieve similar images from the satellite imagery dataset.

\subsubsection{Zero-Shot Image Retrieval}
\begin{figure*}[h]
    \centering
    \includegraphics[width=1\textwidth]{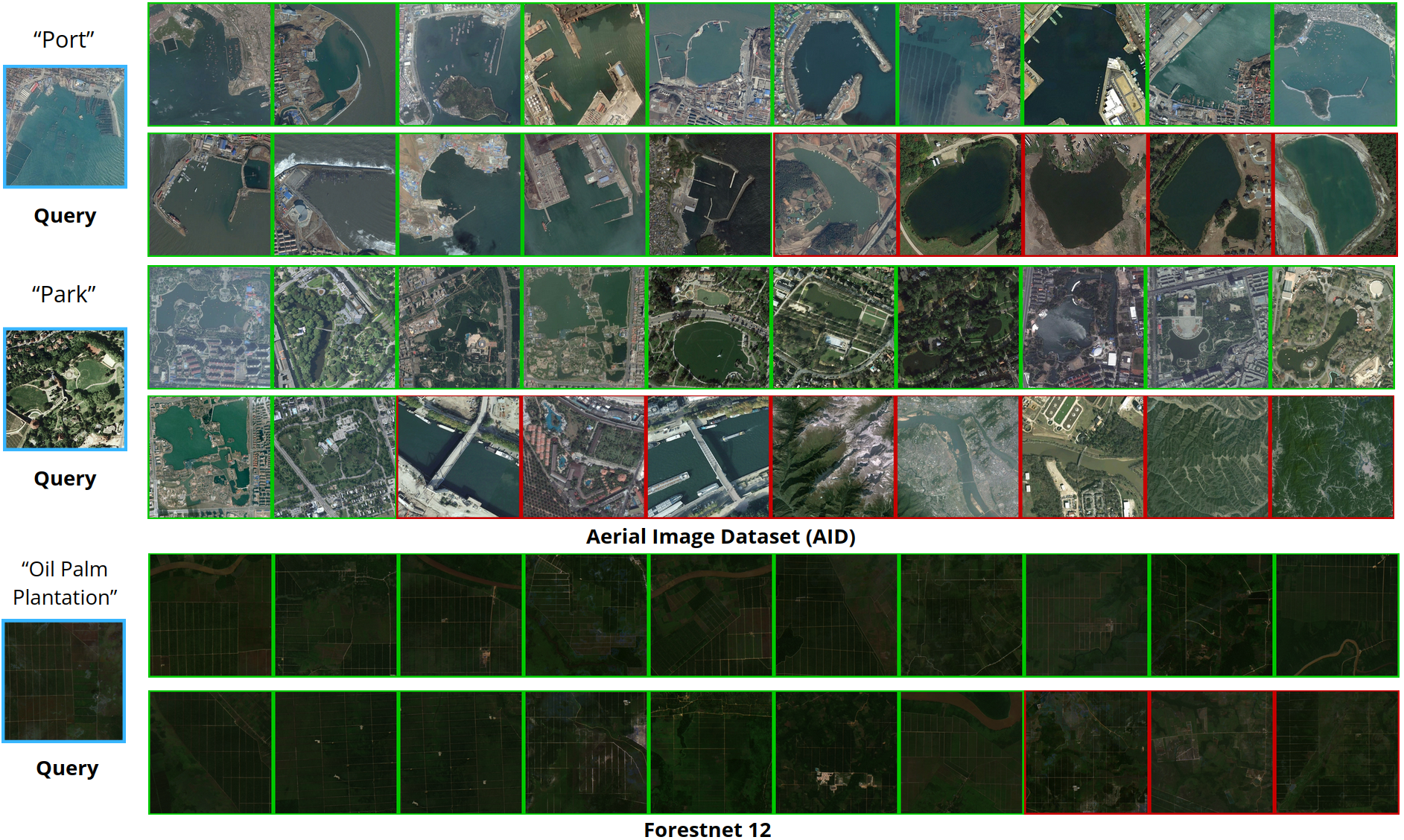}
    \caption{Top 20 retrieved results across the AID and Forestnet datasets. Correct matches are delineated by green frames and incorrect matches by red ones.}
    \label{fig:retrieval_results}
\end{figure*}
The goal of zero-shot image retrieval is to accurately retrieve semantically similar scenes from a large, unindexed database given a query image. The datasets used to evaluate the method are detailed below.

The \textbf{Aerial Image Dataset (AID)} \cite{aid} is a large-scale dataset for aerial scene classification. We evaluate image retrieval on this dataset due to the presence of scenes that are compositional in nature. For example, a harbor scene is composed not only of the water body but also surrounding ships, docks, and piers. AID comprises 10,000 images across 30 classes, ranging from dense urban areas to natural landscapes. 
To assess generalization, we adopt a cross-validation strategy, splitting the data into 5 folds. 
We also use the \textbf{ForestNet dataset} \cite{forestnet} for this task. This dataset was originally designed to classify drivers of deforestation in the Amazon basin. We employ it here to evaluate the model's ability to retrieve fine-grained land-cover mixtures. ForestNet imagery captures complex transitions between primary forest, smallholder agriculture, infrastructure, and commercial clearing. We evaluate the proposed composition-aware approach on both the fine-grained 12-class variant and the broader 4-superclass variant. A single cell often contains a mixture of intact forest and various stages of clearing, so this dataset serves as a rigorous testbed for composition-aware embeddings.

\begin{table}[h]
\centering
\caption{Retrieval Performance (Recall@$K$ and mAP@$K$) on AID and ForestNet Datasets}
\label{tab:cbir_results}
\resizebox{\columnwidth}{!}{%
\begin{tabular}{l c cccccc}
\toprule
\textbf{Model} & \textbf{Params} & \multicolumn{2}{c}{\textbf{@1}} & \multicolumn{2}{c}{\textbf{@5}} & \multicolumn{2}{c}{\textbf{@10}} \\
\cmidrule(lr){3-4} \cmidrule(lr){5-6} \cmidrule(lr){7-8}
& & R & mAP & R & mAP & R & mAP \\
\midrule

\multicolumn{8}{c}{\textit{AID Dataset (5-Fold Cross Validation)}} \\
\midrule
DynamicVis-B \cite{chen2024dynamicvis} & 36.8M & 0.653 & 0.653 & 0.854 & 0.704 & 0.911 & 0.670 \\ 
DynamicVis-L \cite{chen2024dynamicvis} & 91.3M & 0.699 & 0.699 & 0.887 & 0.746 & 0.936 & 0.713 \\ 
SatMAE \cite{cong2022satmae} & 303M & 0.786 & 0.786 & 0.930 & 0.819 & 0.962 & 0.782 \\ 
Prithvi-EO-2.0 \cite{prithvi} & 600M & 0.625 & 0.625 & 0.841 & 0.676 & 0.907 & 0.638 \\ 
\textbf{Proposed Method} & 36.8M & \textbf{0.802} & \textbf{0.802} & \textbf{0.934} & \textbf{0.831} & \textbf{0.960} & \textbf{0.808} \\

\midrule
\multicolumn{8}{c}{\textit{ForestNet-12 (Fine-Grained)}} \\
\midrule
DynamicVis-B \cite{chen2024dynamicvis} & 36.8M & 0.148 & 0.148 & 0.509 & 0.267 & 0.759 & 0.279 \\
DynamicVis-L \cite{chen2024dynamicvis} & 91.3M & 0.152 & 0.152 & 0.543 & 0.279 & 0.732 & 0.282 \\
SatMAE \cite{cong2022satmae} & 303M & 0.317 & 0.317 & 0.639 & 0.419 & 0.803 & 0.410 \\ 
Prithvi-EO-2.0 \cite{prithvi} & 600M & 0.321 & 0.321 & \textbf{0.706} & 0.433 & \textbf{0.852} & 0.427 \\ 
\textbf{Proposed Method} & 36.8M & \textbf{0.347} & \textbf{0.347} & 0.683 & \textbf{0.447} & 0.833 & \textbf{0.434} \\

\midrule
\multicolumn{8}{c}{\textit{ForestNet-4 (Superclass)}} \\
\midrule
DynamicVis-B \cite{chen2024dynamicvis} & 36.8M & 0.332 & 0.332 & 0.799 & 0.489 & 0.942 & 0.467 \\
DynamicVis-L \cite{chen2024dynamicvis} & 91.3M & 0.351 & 0.351 & 0.820 & 0.495 & 0.949 & 0.475 \\
SatMAE \cite{cong2022satmae} & 303M & 0.501 & 0.501 & 0.867 & 0.617 & 0.954 & 0.589 \\ 
Prithvi-EO-2.0 \cite{prithvi} & 600M & 0.507 & 0.507 & \textbf{0.896} & 0.616 & \textbf{0.961} & 0.590\\ 
\textbf{Proposed Method} & 36.8M & \textbf{0.579} & \textbf{0.579} & 0.892 & \textbf{0.666} & 0.956 & \textbf{0.621} \\

\bottomrule
\end{tabular}%
}
\end{table}
As shown in Table \ref{tab:cbir_results}, our composition-aware pretraining yields significant performance gains across all retrieval cutoffs (@$K$). On the AID dataset, we observe a 20.5\% relative improvement in mAP@10 (0.670 to 0.808). The performance gap is even more pronounced on the highly compositional ForestNet dataset. For the fine-grained ForestNet-12 variant, our method almost doubles the baseline mAP@10 (0.279 to 0.434), proving that explicitly modeling fractional land-cover significantly enhances the model's ability to distinguish subtle differences in complex environments.

Qualitative retrieval results (Figure \ref{fig:retrieval_results}) demonstrate the success of the composition-aware pretraining approach. 
We observe excellent retrieval performance on the AID dataset. The visualised retrieval examples demonstrate the model sucessfully learnt the compositional makeup for the "Port" and "Park" classes, evidenced by more than half of the retrieved results being relevant. 
We observe similar results on the fine-grained ForestNet dataset; in the example shown, the model was able to sucessfully distinguish the specifics of an "Oil Palm Plantation" from regular ground-cover by modeling the compositional makeup.

\subsubsection{Scene Classification}
This task requires assigning a single label to an entire aerial or satellite image patch based on its dominant land-cover characteristics. This task evaluates the model's capacity for global semantic understanding and its ability to aggregate complex local textures into a single categorical representation.

We evaluate whole-image semantic understanding using two datasets. The first is the \textbf{UC Merced Land Use dataset} \cite{uc-merced}. This benchmark contains 2,100 high-resolution aerial images extracted from the USGS National Map Urban Area Imagery collection. The dataset is divided into 21 specific land-use classes, with exactly 100 images per class. The second dataset we use is the \textbf{NWPU RESISC45 dataset} \cite{nwpu}. It contains 31,500 images with 45 scene categories. For this task, we extracted pooled features from the frozen backbones and trained a simple Multi-Layer Perceptron (MLP) head to predict the class label.

Despite the presence of highly overlapping visual features between classes (such as ``freeway'' vs. ``intersection'') on both datasets, our composition-aware backbone outperforms the DynamicVis-B baseline. On the UC Merced dataset, our method achieves an Accuracy of 0.928, significantly outperforming the DynamicVis baseline (0.864), as shown in Table \ref{tab:sc_results}. The bidirectional MIL alignment during pretraining ensures that these global scene representations remain categorically separated, while the compositional histogram awareness provides highly discriminative features for accurate class determinations.

On the NWPU-RESISC45 dataset, the proposed method achieves 81.60\% accuracy, which represents a substantial improvement over the DynamicVis-B baseline (0.692) but falls marginally below the SatMAE (0.829) model.

\begin{table}[h]
\centering
\caption{Scene Classification Performance on the UC Merced and NWPU-RESISC45 Datasets}
\label{tab:sc_results}
\resizebox{\columnwidth}{!}{%
\begin{tabular}{l c cccc}
\toprule
\textbf{Model} & \textbf{Params} & \textbf{Accuracy} & \textbf{Precision} & \textbf{Recall} & \textbf{F1 Score} \\
\midrule

\multicolumn{6}{c}{\textit{UC Merced Dataset}} \\
\midrule
DynamicVis-B \cite{chen2024dynamicvis} & 36.8M & 0.864 & 0.866 & 0.864 & 0.863 \\
DynamicVis-L \cite{chen2024dynamicvis} & 91.3M & 0.864 & 0.866 & 0.864 & 0.863 \\
Prithvi-EO-2.0 \cite{prithvi} & 600M & 0.881 & 0.885 & 0.881 & 0.882 \\
SatMAE & 303M \cite{cong2022satmae} & 0.902 & 0.903 & 0.902 & 0.901 \\
\textbf{Proposed Method} & 36.8M & \textbf{0.928} & \textbf{0.930} & \textbf{0.928} & \textbf{0.927} \\

\midrule
\multicolumn{6}{c}{\textit{NWPU-RESISC45 Dataset}} \\
\midrule
DynamicVis-B \cite{chen2024dynamicvis} & 36.8M & 0.692 & 0.700 & 0.692 & 0.692 \\
DynamicVis-L \cite{chen2024dynamicvis} & 91.3M & 0.776 & 0.779 & 0.776 & 0.776 \\
Prithvi-EO-2.0 \cite{prithvi} & 600M & 0.807 & 0.811 & 0.807 & 0.807 \\
SatMAE \cite{cong2022satmae} & 303M & \textbf{0.829} & \textbf{0.834} & \textbf{0.829} & \textbf{0.829} \\
\textbf{Proposed Method} & 36.8M & 0.816 & 0.818 & 0.816 & 0.815 \\

\bottomrule
\end{tabular}%
}
\end{table}
\subsection{Evaluation on Instance-Level Perception and Pixel-Level Dense Prediction}

To assess the versatility of the learned representations, we evaluate our model on dense prediction and instance-level tasks, specifically change detection, tiny object detection, and semantic segmentation. While our composition-aware pretraining explicitly optimizes for region-level semantic mixtures, it remains critical that the foundation model preserves localized spatial features for exact pixel-level tasks. A Feature Pyramid Network is coupled with a Faster R-CNN head for instance detection \cite{ren2016fasterrcnnrealtimeobject}.
Pixel-level Interpretation targets dense prediction tasks, such as change detection and segmentation. The feature differences between bi-temporal inputs are computed in high-dimensional semantic space, followed by an MLP head to produce the final change prediction. We compare our approach with only Dynamicvis-B and L on these tasks due to the presence of an FPN that performs top-down cross-scale feature fusion. This mechanism reconstructs multi-scale feature maps with a uniform channel dimension. This ensures compatibility with standard prediction heads, such as the Mask R-CNN \cite{ren2016fasterrcnnrealtimeobject} and UperNet\cite{upernet}.

\subsubsection{Change Detection}
\begin{figure}[h]
    \centering
    \includegraphics[width=\linewidth]{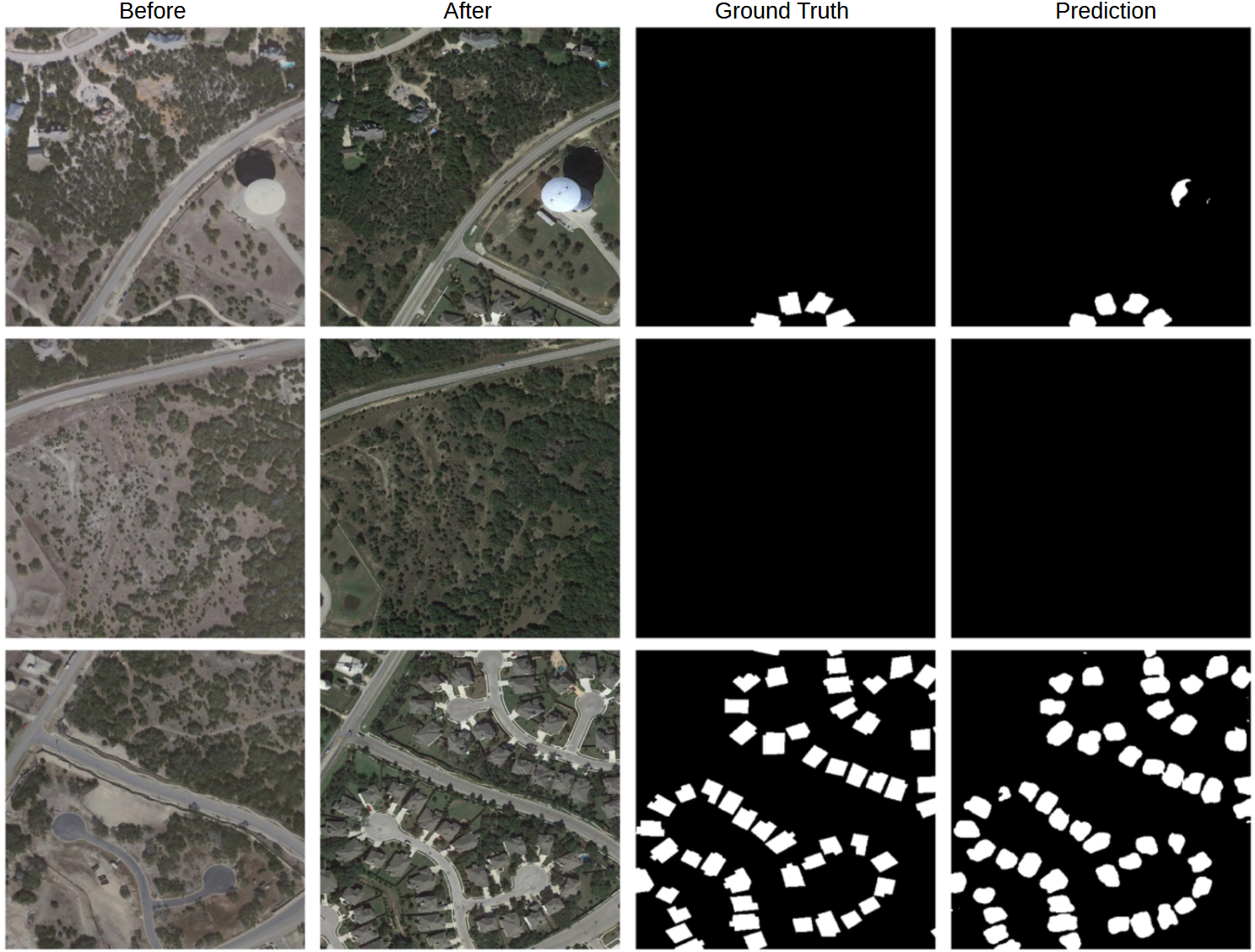}
    \caption{Qualitative results on the LEVIR-CD dataset. From left to right: Pre-event image, Post-event image, Ground Truth mask, and Model Prediction.}
    \label{fig:cd_sample}
\end{figure}
Change detection is the process of identifying meaningful structural or land-cover modifications across bi-temporal image pairs of the same geographic region. This task requires the FM to isolate localized physical changes while remaining invariant to superficial differences such as seasonal variations, atmospheric conditions, or lighting. For the downstream task of change detection, we utilize the \textbf{LEVIR-CD (LEVIR Change Detection) dataset} \cite{levir-cd}. LEVIR-CD is a large-scale dataset specifically curated for identifying building construction and demolition. It consists of over 600 bi-temporal high-resolution Google Earth images, capturing the same geographical regions across temporal spans of 5 to 14 years. The dataset provides pixel-level binary annotations. On the LEVIR-CD dataset, the foundation model must identify structural changes across bi-temporal image pairs. To evaluate this, we trained a lightweight prediction head for 50 epochs while keeping the respective backbones strictly frozen. 

As detailed in Table \ref{tab:cd_results}, our composition-aware method achieves an F1-score of 0.806 and an Intersection over Union (IoU) of 0.680. While this demonstrates robust sensitivity to fine-grained structural evolution, the vanilla DynamicVis achieves slightly higher spatial precision on this specific task (F1: 0.805, IoU: 0.690). Figure \ref{fig:cd_sample} illustrates a sample prediction, highlighting the model's ability to localize modifications. The marginal gap relative to the vanilla DynamicVis-B baseline shows a trade-off inherent to composition-aware pretraining. The composition objective trains the backbone to encode fractional distributions over entire cells, reinforcing holistic mixture representations at the expense of the sharp, high-frequency spatial gradients that precise boundary localization demands. Consequently, the backbone learns to subordinate precise boundary information in favor of semantically richer, cell-level compositional summaries. This trade-off yields significant gains on region-level tasks while incurring a modest cost on pixel-level ones.

\begin{table}[h]
\centering
\caption{Change Detection Performance on the LEVIR-CD Dataset. The proposed framework maintains highly competitive structural sensitivity using only 36.8M parameters.}
\label{tab:cd_results}
\resizebox{\columnwidth}{!}{%
\begin{tabular}{l c cccc}
\toprule
\textbf{Model} & \textbf{Params} & \textbf{Precision} & \textbf{Recall} & \textbf{F1 Score} & \textbf{IoU} \\
\midrule
DynamicVis-B \cite{chen2024dynamicvis} & 36.8M & 0.813 & 0.780 & 0.805 & 0.690 \\
DynamicVis-L \cite{chen2024dynamicvis} & 91.3M & \textbf{0.838} & \textbf{0.824} & \textbf{0.830} & \textbf{0.712} \\
\textbf{Proposed Method} & 36.8M & 0.801 & 0.822 & 0.806 & 0.680 \\
\bottomrule
\end{tabular}%
}
\end{table}

\subsubsection{Tiny Ship Detection}
Tiny object detection involves localizing and classifying extremely small targets within expansive backgrounds. In remote sensing, this task evaluates whether a foundation model preserves high-resolution, fine-grained spatial activations rather than washing them out through pooling operations, which is critical for identifying minimal objects like vessels or vehicles.

The \textbf{LEVIR-Ship dataset} \cite{levir-ship} presents a severe challenge for foundation models, as tiny objects occupy only a fraction of the spatial resolution. This dataset is designed specifically for tiny object detection in optical remote sensing imagery. It features thousands of high-resolution scenes with heavily clustered and multi-scale ship instances. Because the targets often occupy only a few pixels relative to a massive background of water or port infrastructure, this task heavily relies on the foundation model's ability to preserve high-resolution, localized spatial features rather than collapsing them into a holistic background embedding. To assess localization capabilities, we trained a detection decoder for 50 epochs over the frozen backbones. As reported in Table \ref{tab:ship_results}, our composition-aware model achieves an mAP@0.50 of 0.619. While the model successfully localizes targets, the proposed approach demonstrates a stronger performance across both IoU thresholds compared to the DynamicVis baseline but is consistently outperformed by the large variant.


\begin{table}[h]
\centering
\caption{Tiny Ship Detection Performance on the LEVIR-Ship Dataset.}
\label{tab:ship_results}
\resizebox{\columnwidth}{!}{%
\begin{tabular}{l c ccccc}
\toprule
\textbf{Model} & \textbf{Params} & \textbf{mAP@0.10} & \textbf{mAP@0.50} & \textbf{Precision} & \textbf{Recall} \\
\midrule
DynamicVis-B \cite{chen2024dynamicvis} & 36.8M & 0.730 & 0.552 & 0.135 & 0.870 \\
DynamicVis-L \cite{chen2024dynamicvis} & 91.3M & \textbf{0.759} & \textbf{0.643} & \textbf{0.162} & \textbf{0.904} \\
\textbf{Proposed Method} & 36.8M & 0.742 & 0.619 & 0.140 & 0.875 \\
\bottomrule
\end{tabular}%
    }
\end{table}

\subsubsection{Semantic Segmentation}
\begin{figure}
    \centering
    \includegraphics[width=\linewidth]{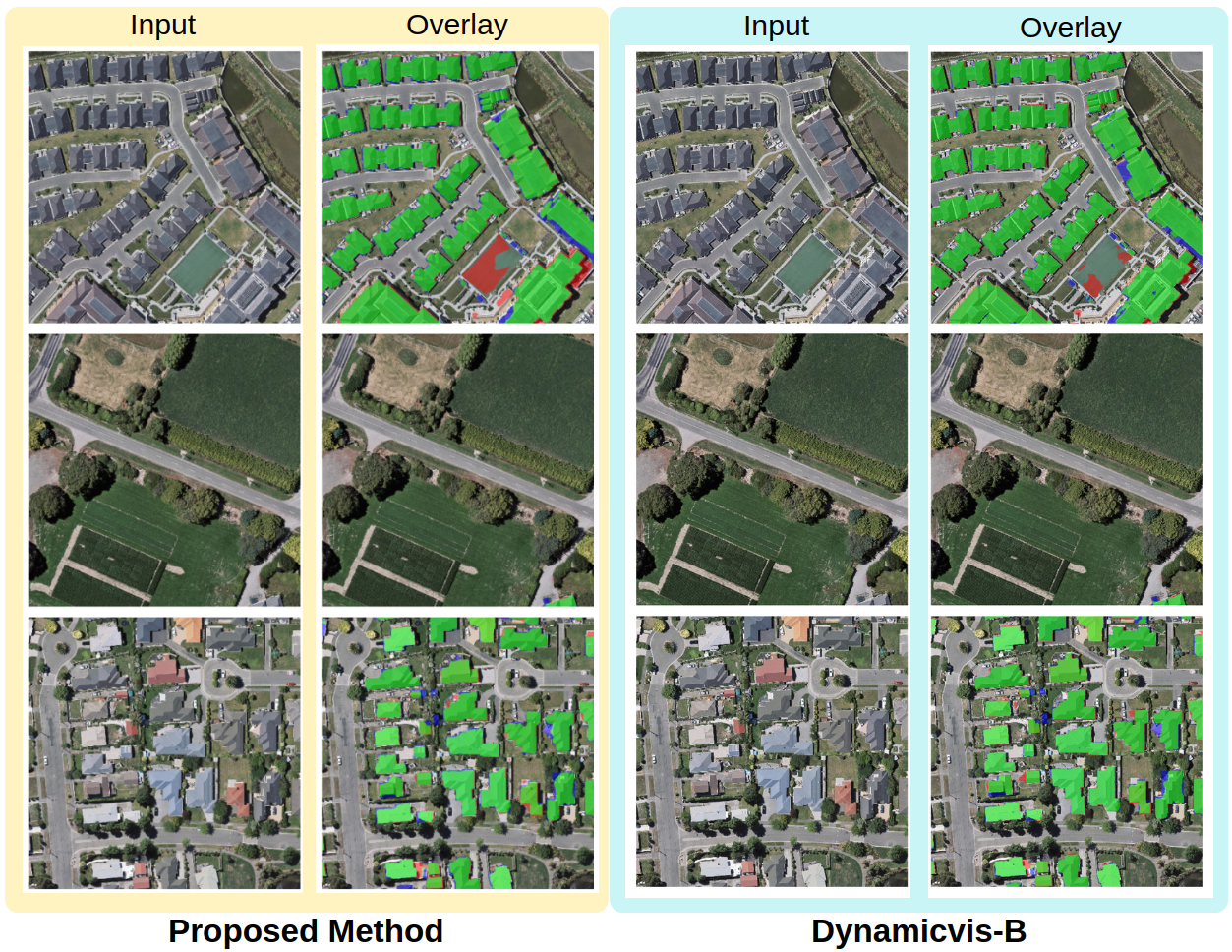}
    \caption{Qualitative results on the WHU building dataset show that the model successfully learnt to delineate structures. The overlay demonstrates the true positives in green, false negatives in blue, and false positives in red.}
    \label{fig:whu-dataset}
\end{figure}

\begin{table}[h]
\centering
\caption{Semantic Segmentation Performance on the WHU Building Dataset. Our method delivers a competitive IoU while baseline preserves slightly better sub-pixel boundary delineations due to its lack of cell-level pooling constraints.}
\label{tab:seg_perf}
\resizebox{\columnwidth}{!}{%
\begin{tabular}{l c cccc}
\toprule
\textbf{Model} & \textbf{Params} & \textbf{Precision} & \textbf{Recall} & \textbf{F1 Score} & \textbf{IoU} \\
\midrule
DynamicVis-B \cite{chen2024dynamicvis} & 36.8M & 0.936 & 0.914 & 0.925 & 0.860 \\
DynamicVis-L \cite{chen2024dynamicvis} & 91.3M & \textbf{0.964} & \textbf{0.941} & \textbf{0.953} & \textbf{0.886} \\
\textbf{Proposed Method} & 36.8M & 0.934 & 0.908 & 0.921 & 0.853 \\
\bottomrule
\end{tabular}%
}
\end{table}

We evaluate semantic segmentation performance on the single-category \textbf{WHU building dataset}~\cite{whu-building-dataset}, comprising discrete, localized instance-like structures. UperNet~\cite{upernet} is used as the prediction framework. It couples a Feature Pyramid Network with a Unified Perceptual Parsing head that aggregates multi-scale feature maps from the frozen backbone across hierarchical stages, enabling dense per-pixel predictions without requiring any modification to the backbone architecture. We observe that the model does learn to successfully delineate structures, but it falls slightly short of the DynamicVis-B baseline. As visible in Figure \ref{fig:whu-dataset} and Table \ref{tab:seg_perf}, the DynamicVis baseline produces fewer false positives, consistent with its stronger preservation of fine-grained spatial topology. This result corroborates the resolution-semantics trade-off identified earlier. The composition-aware pretraining framework optimizes representations for distributional similarity over entire cells, and this bias, while highly effective for semantic retrieval and scene classification, modestly reduces the sub-pixel boundary precision that dense prediction tasks require.

\subsection{Architecture Analysis}
We perform ablation studies on the number of vocabulary classes and different loss configurations. The models were trained with half the training budget of the final models reported in the earlier results. 
\subsubsection{Vocabulary Size}

The visual vocabulary size $k$ governs the granularity of the composition targets. A vocabulary that's too small collapses semantically distinct textures into the same centroid, producing low-variance targets that provide a weak training signal. An excessively large vocabulary over-partitions the embedding space and creates sparse distributions.

To determine the optimal vocabulary size, we trained the full pretraining pipeline with $k \in \{32, 64, 128, 256, 512, 1024\}$ and evaluated each configuration on the zero-shot image retrieval task with the AID dataset. Table~\ref{tab:vocab_ablation} reports the full retrieval metrics with standard deviations across folds.

\begin{table}[h]
\caption{Vocabulary size ablation on the AID dataset.}
\label{tab:vocab_ablation}
\small
\setlength{\tabcolsep}{4pt}
\begin{tabular}{lcccccc}
\toprule
$k$ & R@1 & mAP@1 & R@5 & mAP@5 & R@10 & mAP@10 \\
\midrule
32   & 0.735\tiny{±.007} & 0.735\tiny{±.007} & 0.902\tiny{±.002} & 0.774\tiny{±.002} & 0.943\tiny{±.002} & 0.740\tiny{±.002} \\
64   & 0.727\tiny{±.009} & 0.727\tiny{±.009} & 0.914\tiny{±.006} & 0.774\tiny{±.007} & 0.950\tiny{±.003} & 0.738\tiny{±.006} \\
128  & 0.744\tiny{±.008} & 0.744\tiny{±.008} & 0.913\tiny{±.006} & 0.784\tiny{±.006} & 0.953\tiny{±.006} & 0.750\tiny{±.006} \\
256  & 0.734\tiny{±.008} & 0.734\tiny{±.008} & 0.906\tiny{±.009} & 0.774\tiny{±.008} & 0.946\tiny{±.009} & 0.739\tiny{±.008} \\
\textbf{512}  & \textbf{0.751}\tiny{±.013} & \textbf{0.751}\tiny{±.013} & \textbf{0.919}\tiny{±.011} & \textbf{0.794}\tiny{±.010} & \textbf{0.952}\tiny{±.009} & \textbf{0.761}\tiny{±.008} \\
1024 & 0.707\tiny{±.012} & 0.707\tiny{±.012} & 0.900\tiny{±.008} & 0.756\tiny{±.008} & 0.943\tiny{±.004} & 0.721\tiny{±.006} \\
\bottomrule
\end{tabular}
\end{table}

$k=512$ achieves the highest mAP@10 of $0.761 \pm 0.008$, outperforming both smaller and larger vocabularies. The performance drop at $k=1024$ is the most pronounced, with a $-0.040$ reduction in mAP@10 relative to $k=512$. We hypothesize that this degradation is due to vocabulary over-partitioning. The clear advantage of $k=512$ over this range is the primary basis for our vocabulary size selection.

\subsubsection{Loss Formulation Ablation}
\label{sec:loss_ablation}
To validate each component of the multi-branch training objective, we conduct a systematic ablation study over subsets of the three loss
terms. The relative weighting of $\lambda_{mil}$ and $\lambda_{cls}$ preserves the 
ratio established in the DynamicVis pretraining objective~\cite{chen2024dynamicvis}. 
The compositional distillation weight $\lambda_{emd}$ is then set at a 2:1 
ratio to $\lambda_{mil}$. Seven configurations are evaluated on AID mAP@10 and ForestNet-4 mAP@10. Table~\ref{tab:loss_ablation} summarises the results.

\begin{table}[h]
\caption{Loss component ablation on AID and ForestNet-4.}
\label{tab:loss_ablation}
\begin{tabular}{lcc}
\toprule
Configuration & AID mAP@10 & FN-4 mAP@10 \\
\midrule
EMD only              & 0.750\tiny{±.007} & 0.598 \\
EMD + CLS             & 0.805\tiny{±.006} & 0.617 \\
EMD + MIL             & 0.759\tiny{±.006} & 0.578 \\
EMD + MIL + CLS       & \textbf{0.812}\tiny{±.004} & \textbf{0.621} \\
\midrule
CLS only              & 0.646\tiny{±.006} & 0.516 \\
MIL only              & 0.611\tiny{±.005} & 0.572 \\
MIL + CLS             & 0.719\tiny{±.008} & 0.476 \\
\bottomrule
\end{tabular}
\end{table}
\noindent\textbf{The EMD loss is the primary driver of representation quality.}
The performance gap between EMD-inclusive and EMD-absent configurations is substantial and consistent across both datasets. Removing EMD entirely while retaining MIL and CLS yields a drop of $-0.148$ in AID mAP@10. This confirms that the histogram distillation objective is the core source of compositional discriminability in the learned
representations, and that the auxiliary branches cannot substitute for it.


\section{Conclusions and Future Work}

This paper introduces a composition-aware pretraining framework that resolves the single-concept bottleneck in modern geospatial foundation models. This is achieved by explicitly encoding the heterogeneous land-cover mixtures fundamental to Earth observation data. Evaluations across region-level understanding, instance-level perception, and pixel-level dense prediction highlight a distinct representational trade-off. The EMD objective trains the backbone to encode distributional summaries over entire cells, naturally subordinating precise boundary information in favor of holistic mixture representations. As a result, this composition-aware prior dramatically benefits tasks requiring region-level understanding, such as zero-shot image retrieval and scene classification, while incurring only a minimal performance cost on sub-pixel tasks like change detection and semantic segmentation.

The strength of explicitly modeling fractional composition is most evident in zero-shot image retrieval. On the fine-grained ForestNet-12 dataset, designed to capture subtle transitions between intact forests and agricultural clearing, our method almost doubles the DynamicVis-B baseline mAP@10 (0.279 to 0.434). By forcing the model to represent not only which land-cover types are present, but in what exact proportions, the framework develops the representational capacity required to distinguish highly nuanced scenes, such as separating an oil palm plantation from adjacent secondary growth. Furthermore, on the Aerial Image Dataset (AID), our method yields a 20.5\% relative improvement in mAP@10, jumping from 0.670 to 0.808. The results on tasks requiring instance and pixel level prediction validate that although the model sacrifices a marginal degree of sub-pixel boundary precision, it successfully preserves the localized spatial features necessary for these tasks. 

There are several promising directions for further exploration. First, the pretraining methodology could be extended to a broader range of architectures, including standard Vision Transformers and modern CNNs, to establish its universality beyond the DynamicVis state-space model. Second, because the current framework strictly processes 3-channel RGB imagery, expanding the formulation to accommodate multispectral and hyperspectral data could unlock even richer compositional insights. Third, replacing the current offline K-Means clustering with end-to-end learnable vocabularies would allow the model to dynamically discover task-relevant textures during pretraining, rather than relying on static initialization. Finally, to address the resolution-semantics trade-off observed in our pixel-level evaluations, applying histogram targets at multiple spatial resolutions simultaneously could recover fine-grained spatial precision. This would bridge the performance gap in dense prediction tasks without compromising the highly discriminative region-level representations that define this framework.

\bibliographystyle{ACM-Reference-Format}
\bibliography{sample-base}










\end{document}